\documentclass[ijoc,nonblindrev]{informs3} 

\OneAndAHalfSpacedXII 

\usepackage{natbib}
 \bibpunct[, ]{(}{)}{,}{a}{}{,}%
 \def\bibfont{\small}%
\usepackage{algorithm2e}
\usepackage{algorithmic}
\usepackage{mathtools}
\usepackage{subfigure}
\newcommand{\Norm}[1]{\left\lVert#1\right\rVert}
\newtheorem{lem}{Lemma}
\newtheorem{thm}{Theorem}
\TheoremsNumberedThrough     

\EquationsNumberedThrough    

\begin{document}




\TITLE{GADGET SVM: A Gossip-bAseD sub-GradiEnT Solver for Linear SVMs}
\ARTICLEAUTHORS{%
\AUTHOR{Haimonti Dutta}
\AFF{Department of Management Science and Systems,\\ State University of New York,\\ Buffalo, NY, 14260 USA \\  \EMAIL{haimonti@buffalo.edu}, \URL{}}
\AUTHOR{Nitin Nataraj}
\AFF{Department of Computer Science, \\State University of New York,\\ Buffalo, NY, 14260 USA \\  \EMAIL{nitinnat@buffalo.edu} \URL{}}
} 

\ABSTRACT{%
\noindent In the era of big data, an important weapon in a machine learning researcher's arsenal is a \emph{scalable} Support Vector Machine (SVM) algorithm. 
SVMs are extensively used for solving classification problems. Traditional algorithms for learning SVMs often scale super linearly with training set size which becomes infeasible very quickly for large data sets. In recent years, scalable algorithms have been designed which study the primal or dual formulations of the problem. This often suggests a way to decompose the problem and facilitate development of distributed algorithms. In this paper, we present a distributed algorithm for learning linear Support Vector Machines in the primal form for binary classification called Gossip-bAseD sub-GradiEnT (GADGET) SVM. The algorithm is designed such that it can be executed locally on nodes of a distributed system; each node processes its local homogeneously partitioned data and learns a primal SVM model; it then gossips with random neighbors about the classifier learnt and uses this information to update the model. Extensive theoretical and empirical results suggest that this anytime algorithm has performance comparable to its centralized and online counterparts.

}%


\HISTORY{}

\maketitle

%


\section{Introduction}
\label{intro}
The evolution of large and complex collections of digital data has necessitated the development of scalable machine learning algorithms (\citet{Rajaraman_11a,Bottou_07a,Bekkerman_11a}). 
These algorithms rely significantly on well established techniques of parallelization and distributed computing (\citet{KC:2000,Zaki_00,Tanenbaum_06, Lynch_96, Bertsekas_97}). Parallel systems for machine learning are often tightly coupled including shared memory systems (SMP), distributed memory machines (DMM) or clusters of SMP workstations (CLUMPS) with fast interconnection between them. Distributed systems, in contrast, are loosely-coupled -- for example, mobile ad-hoc networks or sensor networks (\citet{Bliman_08a, Blondel_05a, Boyd_05a, Cao_08a, Cao_08b, Cao_05p, Carli_07a,Carli_06a, Jadbabaie_03a, Kar_07a,Kashyap_07a,Olfati_04a, Olshevsky_06a}).  


Distributed systems can function without a central server for co-ordination. They are often subject to abrupt changes in topology due to nodes joining or leaving, and are susceptible to link failures. They are collectively capable of storing large amounts of data of different modalities (such as text, audio, video). The data, distributed in the network, can be made highly available by replication. 
Often, distributed machine learning algorithms are designed to execute on data distributed in the network. Such algorithms are extensively used in recent years -- for example,
in distributed sensor networks for co-ordination and control of Unmanned Aerial Vehicles (UAVs), fleet, self-driving cars (\citet{Tortonesi_12, Kargupta_10}), automated products and parts in transportation, life science and energy markets. 

\noindent \textbf{Consensus-based} learning algorithms (\citet{Datta_06a}) are a special class of distributed machine learning algorithms which builds on the seminal work of Tsitsiklis (\citet{Tsitsiklis_84,Bertsekas_97}). Specifically, these algorithms focus on solving separable minimization problems of the form:  minimize $F(x) = \sum_{i=1}^n f_i(x) $, subject to $x \in X$, where X is a convex constraint set; $n$ is the number of nodes in the network, $F(x)$ is the global function to be learnt and $f_i(x)$ are local to the nodes. It is typically assumed that the functions $f_i(x)$ are Lipschitz continuous, possibly with Lipschitz continuous gradient. Related literature (\citet{Cao_08a,Cao_08b,Cao_05p}) also focuses on distributing computations involved with optimizing a global objective
function among different processors (assuming complete information about the global objective function at each processor). They have also been studied in the context of multi-agent systems where they are used for ensuring cooperative behavior among agents. These algorithms assume that the individual values of an agent can be processed and are unconstrained. In recent times, they have been studied from a game-theoretic perspective (\citet{Lin_15a,Lin_14a,Marden_07a}). In this approach, the agents are endowed with local utility functions that lead to a game form with a Nash equilibrium which is the same as or close to a global optimum. Various learning algorithms can then be designed to help agents reach the equilibrium. 

The following are some characteristics exhibited by these algorithms: (1) they are completely decentralized (2) rely on local observations, but capable of making global inferences by communicating essential information with neighbors; (3) completely asynchronous; (4) resilient to changes in underlying topology, and (5) scalable in the size of the network. They are capable of learning \emph{cooperatively} and when in agreement, can reach a \emph{consensus}. At that time, the nodes have a good approximation\footnote{$\epsilon$ tolerance, where $\epsilon$ is usually user-defined.} of the global solution.

In this paper, we present a consensus based linear Support Vector Machine algorithm for binary classification called \textbf{G}ossip-b\textbf{A}se\textbf{D} sub-\textbf{G}radi\textbf{E}n\textbf{T}  solver (GADGET SVM). We assume that there is a network of computational units, each containing horizontally\footnote{Horizontal partitioning implies that each node has the same set of features or attributes.} partitioned samples of data instances. Nodes are capable of building support vector machine models from  \emph{local} data. They can update local models by exchanging information with their neighbors. The overall goal, is to learn at each node, a close approximation of the \emph{global} function. Communication between nodes is permitted by use of a \emph{gossip}-based protocol i.e. each node contacts a neighbor at random and exchanges information. The process continues until there are no significant changes\footnote{measured by a user defined $\epsilon$ parameter} in the local weight vector. This algorithm is an \emph{anytime} algorithm without any predefined termination criteria. 

\noindent{\textbf{Organization }} The rest of this paper is organized as follows: Section~\ref{back} provides the background and related work; 
Section~\ref{prelim} presents the GADGET SVM algorithm and its theoretical foundations; Section~\ref{empirical} presents empirical results on real world data; Section~\ref{conc} discusses directions of future work and concludes the paper.

\section{Background}
\label{back}

\subsection{Support Vector Machines - A brief review}
\label{review}
\noindent Formally, given a training set  $S$ comprising of feature vectors $x_i \in \mathbb{R}^n, i=1,2,\cdots,N$ and binary labels $y_i \in \{-1,+1\}$, the goal is to find a linear classifier of the form $f(x)=w_i^T x_i + b, b \in \mathbb{R}, w \in \mathbb{R}^n$. More generally, in the primal SVM formulation the goal is to find the minimizer of the problem
  
\begin{equation}
\label{DSVM}
\min_{\mathbf{w}} \frac{\lambda}{2} \parallel{\mathbf{w}}^2 \parallel + \frac{1}{N} \sum_{j = 1}^{N} l(\mathbf{w}; (\mathbf{x_j},y_j)),
\end{equation}

where $l$ is the loss function defined as
$l(\mathbf{w}; (\mathbf{x},y)) = \max \{0, 1 - y\langle \mathbf{w}, \mathbf{x} \rangle \}$ for hinge loss 
and $\lambda$ is the SVM regularization parameter. The above is unconstrained and piecewise quadratic and can be written as a convex QP. The dual is also a convex QP in the variable $\alpha=(\alpha_1, \alpha_2, \cdots, \alpha_N)^T$ given by
\begin{equation}
\label{dualQP}
\min_\alpha \frac{1}{2} \alpha^T K \alpha - \textbf{1}^T \alpha \text{ such that } 0 \le \alpha \le C\textbf{1}, y^T\alpha=0 
\end{equation}
where $K_{ij}=(y_iy_j)x_i^Tx_j, y = (y_1, y_2,\cdots, y_N)^T, \textbf{1}=(1, 1, \cdots,1)^T$. The Karush Kuhn Tucker conditions relate the primal and dual solutions by $w=\sum_{i=1}^{N} \alpha_i y_i x_i$ and $b$ is the Lagrange multiplier for $y^T\alpha=0$ in Equation~\ref{linearKKT}. The linear classifier can therefore be written as 
\begin{equation}
\label{linearKKT}
f(x)=\sum_{i=1}^{N} \alpha_i y_i (x_i^T x) + b.
\end{equation}
From a risk minimization perspective, Equation~\ref{DSVM} can also be written as 
\begin{equation}
\label{risk}
\min_{w,b,\xi} \frac{1}{2} \parallel{\mathbf{w}}^2 \parallel + C \sum_{i=1}^{N} \xi_i,
\end{equation}
subject to $\xi_i \ge 0, \text{ } y_i (w^Tx_i + b) \ge 1 - \xi_i, \text{ } i = 1,2,\cdots,N $ where $\xi_i = |y_i - f(x_i)|$ measures the training error. This can also be written as,

\begin{equation}
\label{riskOverall}
\min_{w,b} P(w,b) = \frac{1}{2} \parallel {\mathbf{w}}_{2}^2 \parallel + R(w,b)
\end{equation}
where $R$ is a piecewise linear function. \\

\subsection{Related Work}
\label{related}
The problem of scaling Support Vector Machine (SVM) algorithms have been studied extensively (\citet{Osuna_97a, Osuna_97b, Joachims_98,Menon_09a}) with a majority of the algorithms developing faster variants of the primal, dual or primal-dual formulations. In this section, we first present a recap of existing algorithms for solving the SVM optimization in primal and dual forms. Following this, we present \emph{scalable} SVM algorithms including parallel and distributed variants which are more closely related to the current work.

\subsubsection{Primal Formulations.} Optimizations of the primal formulation of linear SVMs have been studied extensively by \citet{Mangasarian_02} and Keerti et al. (\citet{Keerthi_05a, Keerthi_06a}). \citet{Mangasarian_02} presents finitely terminating Newton methods with (Armijo method) and without the step size parameter and Keerti et al. (\citet{Keerthi_05a}) extend this work by performing exact line searches to determine the step size for $L_2$ loss functions. They also suggest methods to solve the primal SVM formulations using $L_1$ loss by approximating the loss using modified Huber and logistic regression (\citet{Zhang_03a}). \citet{Chapelle_07a} complements the literature by extending the above techniques to the non-linear case. 

Other large scale primal SVM formulations have been solved by using Stochastic Gradient Descent (SGD, \citet{Menon_09a}) methods. Bottou proposed the SVM-SGD (\citet{bottou_11, SVM-SGD}) algorithm which solves Equation \ref{DSVM} and achieves performance comparable to $\text{SVM}^{light}$ and $\text{SVM}^{perf}$ on benchmark datasets. \citet{Zhang_04} studied stochastic gradient descent algorithms on regularized forms of linear prediction methods such as least squares for regression, logistic regression and SVMs for classification. 

One of the popular SGD algorithms, Pegasos (\citet{Shwartz_2007}) operates by choosing a random subset of $k$ training examples, evaluating the sub-gradient of the objective function on these examples and updating the weight vector accordingly. The weight vector is then projected on a ball of radius $\frac{1}{\sqrt \lambda}$. The parameter $k$ does not affect the run time or its convergence to the optimal solution.  

\citet{Duchi_09a} present the FOrward Backward Splitting algorithm (FOBOS) which alternates between two phases - in the first, an unconstrained gradient is estimated. This is followed by solving an instantaneous optimization problem that trades off minimization of the regularization term while keeping close proximity to the result of the first phase.

Cutting-plane methods (\citet{Joachims_09a, Teo_07}) build a piecewise-linear lower-bounding approximation of $R(w,b)$ and Joachims and his colleagues (\citet{Joachims_06, Joachims_09a}) study a specialized formulation for the case of the error rate and call it the ``structural formulation" given by
\vspace{-3.6mm}
\begin{equation}
\label{structural}
\min_{w,\xi \ge 0} P(w,b) = \frac{1}{2} \parallel {\mathbf{w}}_{2}^2 \parallel + C \xi
\end{equation}
s.t. $\forall c \in \{0,1\}^n:\frac{1}{n} w^T \sum_{i=1}^{n} c_i y_i x_i \ge \frac{1}{n} \sum_{i=1}^{n}c_i - \xi$. 

\noindent The above formulation has only one slack variable that is shared across all constraints. Each constraint in this formulation corresponds to the sum of a subset of constraints from Equation~\ref{risk}. The $\frac{1}{n}\sum_{i=1}^{n}c_i$ gives the maximum fraction of training errors possible over each subset and $\xi$ is an upper bound on the fraction of training errors made. To speed up the convergence of the Cutting Plane Algorithm, \citet{franc_08} propose an Optimized Cutting Plane Algorithm for SVMs (OCAS) which aims at optimizing a reduced problem formulated from Equation~\ref{riskOverall} by substituting a piecewise linear approximation for $R$ while leaving the regularization term unchanged. Finally, \citet{Chang_08a} propose a coordinate descent method for solving primal $L_2$-SVM which does not work for the $L_1$ SVM due to its non-differentiability.

To the best of our knowledge, none of the primal SVM formulations discussed above have been used in the context of distributed consensus based learning.

\subsubsection{Dual Formulations.} \citet{Vapnik_95} shows that the training of a Support Vector Machine leads to the following quadratic optimization (QP) problem:

\begin{equation}
\begin{array}{l}
\label{qp}
\text{minimize:   } W(\alpha) = - \sum_{i=1}^{N} \alpha_i + \frac{1}{2} \sum_{i=1}^{N} \sum_{j=1}^{N} y_i y_j \alpha_i \alpha_j K(x_i, x_j) \\
\text{subject to:   } \sum_{i=1}^{N} y_i \alpha_i =0,
\forall i: 0 \le \alpha_i \le C
\end{array}
\end{equation}

\noindent where each component $\alpha_i$ corresponds to the Lagrangian multiplier of $(x_i,y_i)$. Equation~\ref{qp} can be rewritten as:

\begin{equation}
\begin{array}{l}
\text{minimize:   } W(\alpha) = -\alpha^{T}\textbf{1} + \frac{1}{2}\alpha^{T}Q\alpha \\
\text{subject to: } \alpha^{T}y =0, 0\le \alpha \le C\textbf{1}
\end{array}
\end{equation}

\noindent where $(Q)_{ij}= y_i y_j K(x_i, x_j)$ (\citet{Joachims_99a}). The KKT conditions relate the primal and dual forms $w=\sum_{i=1}^{N} \alpha_i y_i x_i$ while $b$ is Lagrange multiplier for $y^{T}\alpha=0$. This leads to the classifier $f(x)=\sum_{i=1}^{N} \alpha_i y_i(x_i^T x_i) + b$. We do not provide a review of scalable techniques for solving the dual SVM formulation -- this being outside the scope of the current work, but an interested reader is referred to \cite{Stolpe_16a} for a detailed overview.

\subsubsection{Distributed and Parallel Support Vector Machines.} 
An overview of algorithms for distributed Support Vector Machines is presented in \citet{Stolpe_16a}. 
Two different kinds of algorithms are discussed here (a) methods designed to run in high performance compute clusters, assuming high bandwidth connections and an unlimited amount of available energy (b) pervasive computing systems (like wireless sensor networks) consisting of battery-powered devices, which usually require algorithms whose primary focus is the preservation of energy.
\citet{SyLiKa1999} proposed a Distributed SVM algorithm which finds support 
vectors locally at each node and then sends them to a central server for processing. This approach 
is promising because it is both highly parallel, and worked on arbitrary topologies; however, 
it did not find the global optimal solution and its communication cost 
depended on the total size of its dataset. The algorithm was improved in \citet{CaCaHo2005} by allowing 
the centralized server to send the support vectors back to the distributed nodes and then repeating 
the process until a global optimum was achieved. Despite reaching optimality, this 
approach was slow due to extensive communication costs. Another approach, Cascade SVM (\citet{Graf_05})  
worked on a top-down network topology and quickly generated a globally optimal solution. In a similar 
vein to the Cascade SVM, \citet{YuRoLi2008} created a DSVM variant suited for Kurtowski graphs. Both algorithms required
specific network topologies and the transfer of support vectors between nodes resulting in large 
communication complexity. DSVM works on arbitrary networks with
a communication complexity independent of the data-set size, and only linear dependence 
on network size. Furthermore, DSVM always works towards optimality and can 
produce an $\epsilon$-accurate solution for any $\epsilon > 0$. \citet{Hazan_08a} present a parallel
algorithm for solving large scale SVMs by dividing the training set amongst a number of processing nodes
each running an SVM sub-problem associated with that training set. The algorithm uses
a parallel (Jacobi) block-update scheme derived from the convex conjugate (Fenchel Duality) form of the original
SVM problem. Each update step consists of a modified SVM solver running in parallel over the sub-problems followed
by a simple global update. The algorithm has a linear convergence rate and takes $O(\log(\frac{1}{\epsilon}))$ iterations to get
$\epsilon$-close to the optimal solution. A distributed block minimization scheme followed by line search was proposed
by \citet{Pechyony_11a} to solve the dual of the linear SVM formulation. Prior work has shown that sequential block minimization can be
used to solve this formulation by considering a block $B_i$ at each iteration and solving only for the variables in $B_i$. In the distributed
block minimization proposed by \citet{Pechyony_11a} all the blocks are processed simultaneously by $k$ slave nodes in a Mapreduce/Hadoop implementation.
Finally, \citet{Sangkyun_12a} present a framework for training SVMs over distributed sensor networks by making use of multiple local kernels and 
explicit approximations to feature mappings induced by them. 

The algorithm closest in spirit to ours is the consensus based Support Vector Machine algorithm proposed by \citet{Forero_10}. The fundamental differences are (1) the Alternating Direction Method of Multipliers DSVM (MoM-DSVM) solves the dual of the linear SVM formulation given by
\begin{equation}
\displaystyle{\min_{\{w_j, b_j, \epsilon_{jn}\}}} \frac{1}{2} \displaystyle{\sum_{j=1}^J} \parallel w_j^2 \parallel + JC \displaystyle{\sum_{j=1}^J \sum_{n=1}^{N_j}} \epsilon_{jn}
\end{equation}\\
s.t. $y_{jn} (w_j^T x_{jn} + b_j) \ge 1 - \epsilon_{jn}, \forall j \in J, n=1, \cdots, N_j$; 
$\epsilon_{jn} \ge 0, \forall j \in J, n=1, \cdots, N_j$; 
$w_j = w_i, b_j = b_i, \forall j \in J, i \in \mathcal{B}_j$. 
whereas GADGET SVM solves the primal formulation similar to the Pegasos algorithm (\citet{Shwartz_2007}) with modifications due to the distributed nature of the problem. (2) GADGET uses stochastic gradient descent for solving the optimization problem while MoM-DSVM relies on the Alternating Direction Method of Multipliers (\citet{Bertsekas_97, Boyd_11}). (3) The underlying protocol used for communication in GADGET is a randomized gossip algorithm whereby each peer exchanges information with \emph{only one} randomly chosen immediate neighbor within a one-hop distance from itself. In contrast, MoM-DSVM broadcasts its current augmented vector $v_j = [w_{j}^T; b_j]^{T} $ thereby having a higher communication cost than the algorithm described here.

Finally, it must be noted that an earlier version of the GADGET SVM algorithm with two calls to the Push-Sum protocol at each node has been presented at a workshop (\citet{Hensel_09c}). This algorithm has been refined considerably and extensive theoretical and empirical contributions are presented in this paper.


\subsection{Communication protocols - Gossip}
\label{gossip}

Gossip based protocols are popular in distributed systems because of their fault tolerant information dissemination (\citet{Boyd_06,Shah_09a,Dimakis_10a,Dimakis_06a,Narayanan_07a}). Dating back to early work in the database community \citet{Demers_87a}, they provide a simple and effective information spreading strategy, in which every node randomly selects one of its neighbors for message exchange during the process of spread of information.  They are more efficient than widely adopted information exchange protocols such as broadcasting and flooding. Gossip can be used for computation of sums, averages, quantiles, random samples and other aggregate functions and probabilistic guarantees of convergence are ascertained for such computations. The problem of \emph{aggregation} was first proposed by \citet{Bawa_03} wherein it is assumed that there is a network of $n$ nodes, each containing a value $x_i$. The goal is to compute the aggregate functions in a decentralized fault tolerant fashion. \citet{Kempe_03} extend this work further by demonstrating that these protocols converge exponentially fast to the correct answer when using uniform gossip. 

\begin{algorithm}
\textbf{Push-Sum}
\begin{algorithmic}
\STATE Each node maintains a sum $s_{t,i}=x_i$ and weight $w_{t,i}=1$.
\STATE 1. Let $\{(\hat{s}_{t-1,i}, \hat{w}_{t-1,i})\}$ be all the pairs sent to node $i$ in round $t-1$. 
\STATE 2. Let $w_{t,i} = \sum_{t-1} \hat{w}_{t-1,i}$ i.e. perform a sum of all weights received by node $i$ in round $t-1$.
\STATE 3. Let $s_{t,i} = \sum_{t-1,i} \hat{s}_{t-1,i}$ i.e. perform a sum of $\hat{s}_{t-1,i}$ currently at node $i$ with those it received in round $t-1$. 
\STATE 4. Choose shares $\alpha_{t,i,j}$ for each $j$ that node $i$ wishes to communicate with.
\STATE 5. Send $(\alpha_{t,i,j} \times s_{t,i},\alpha_{t,i,j} \times w_{t,i}$) to the node $j$ 
\STATE 6. $\frac{s_{t,i}}{w_{t,i}}$ is the current estimate of the average at node $i$ at time $t$.
\end{algorithmic}
\caption{Push-Sum}
\label{PushSum}
\end{algorithm}

They present the Push-Sum algorithm (also presented in Algorithm~\ref{PushSum} for completeness) which is used in this work for communication amongst nodes in the distributed setting for the GADGET SVM algorithm (presented in Section~\ref{prelim}). It operates as follows -- at all times $t$, each node $i$ maintains a sum $s_{t,i}$, initialized to $s_{0,i} = x_i$, and a weight $w_{t,i}$, initialized to $w_{0,i}= 1$. At time 0, it sends the pair $(s_{0,i}, w_{0,i})$ to itself and in each subsequent time step $t$, each node $i$ follows the protocol given as Algorithm~\ref{PushSum} and updates the weight and sum. The algorithm, as presented, helps to estimate the average in the network. A simple extension to protocol Push-Vector where each node holds a vector $v_{t,i}$ instead of a sum $s_{t,i}$ has also been presented in \cite{Kempe_03}.

\section{The GADGET SVM Algorithm}
\label{prelim}


The Gossip bAseD sub-GradiEnT solver for linear SVMs aims to solve Equation~\ref{DSVM} in a decentralized setting. Let $M$ denote an $N \times d$ matrix with real-valued entries.  This matrix 
represents a dataset of $N$ tuples of the form $x_i \in \mathbb{R}^d, 1 \le i \le N$. Assume this dataset has been \emph{horizontally} distributed over $m$ sites $S_1, S_2, \cdots, S_m$ such that site $S_i$ has a data set $M_{i} \subset M, M_{i}: n_i \times d$ and each $x_j \in M_{i}$ is in $\mathbb{R}^d$. Thus, $M = M_1 \cup M_2 \cup \cdots \cup M_m$ denotes the concatenation of the local datasets. The goal is to learn a linear support vector machine on the global data set $M$, by learning local models at the sites and allowing exchange of information among them using a gossip based protocol. In this work, the local models are constructed using the Pegasos algorithm (\cite{Shwartz_2007}). The implicit assumption is that updating a local model with insight from neighbors is likely to be cheaper than transferring data from all the sites to a central server and also prevents creation of a single point of failure in the distributed setting. We note that algorithms with this flavor have been studied in multi-agent systems (\cite{Nedic_2009b, nedic_2010a}) and optimization literature (\cite{Ram_10a}), for general convex optimization problems using gradient descent and projection style optimization algorithms. Our algorithm extends this literature, by explicitly studying Support Vector Machines in the horizontally partitioned setting with theoretical and empirical analysis.\\

\begin{table}[ht]   
\caption{Summary of Notation}
\centering
\begin{tabular}{l l}
\hline
\\
 
$\hat{\mathbf{w}}_i^{(t)}$ & Node $i$'s weight vector at iteration t\\
$\hat{\mathbf{w}}_{i}^{(t+\frac{1}{2})}$ & Node $i$'s approximate network average update at time $t$\\
$\tilde{\mathbf{w}}_i^{(t+\frac{1}{2})}$ & Node $i$'s update of the local weight vector in the direction of descent\\
$\mathbf{w}^{(t)}$ & Network average weight vector\\
$\hat{L}_i^{(t)}$ & Loss at node $i$ using weight vector $\hat{\mathbf{w}}_i^{(t)}$\\
${L}^{(t)}$ & Loss estimated using weight vector ${\mathbf{w}}^{(t)}$\\
$\lambda$ & Learning parameter\\
${\alpha}^{(t)}$ & Learning parameter\\
$B$ & Doubly stochastic transition probability matrix\\
$n_i$ & Number of training examples at site $i$ \\
$N$ & Total number of training examples in the network\\
$m$ & Total number of nodes\\
$d$ & The dimension of weight vector \\ \hline
\end{tabular}
\label{Symbols}
\end{table}

\noindent \textbf{Model of Distributed Computation. } The distributed algorithm evolves over discrete time with respect to a ``global" clock\footnote{Existence of this clock is of interest only for theoretical analysis}. Each site has access to a local clock or no clock at all. Furthermore, each site has its own memory and can perform local computation (such as estimating the local weight vector). It stores $f_i$, which is the estimated local function. Besides its own computation, sites may receive messages from their neighbors which will help in evaluation of the next estimate for the local function. \\

\noindent \textbf{Communication Protocols. } Sites $S_i$ are connected to one another via an underlying communication framework represented by a graph $G (V, E)$, such that each site  $S_i \in \{S_1, S_2, \cdots , S_m\}$ is a vertex and an edge $e_{ij} \in E$ connects sites $S_i$ and $S_j$. Communication delays on the edges in the graph are assumed to be zero. It must be noted that the communication framework is usually expected to be application dependent. In cases where no intuitive framework exists, it may be possible to simply rely on the physical connectivity of the machines, for example, if the sites $S_i$ are part of a large cluster. \\


\begin{algorithm}[h!]
\small
{
\SetKwData{Left}{left}\SetKwData{This}{this}\SetKwData{Up}{up}
\SetKwFunction{Union}{Union}\SetKwFunction{FindCompress}{FindCompress}
\SetKwInOut{Input}{input}\SetKwInOut{Output}{output}

\textbf{GADGET $(\lambda,T,B)$}\\
\KwIn{$M_i: n_i \times d$ matrix with real valued inputs at each site $S_i; 
G (V, E)$ which encapsulates the underlying communication framework ;\ \\
\textbf{Parameters: } $\lambda$; $\mathcal{T}$; $B$ } 
\BlankLine
\textbf{Initialization:} 
$\hat{\mathbf{w}}_i^{(1)}=0$ \; 
 %
\For{t = 1 to $\mathcal{T}$}
 {
  (a) Choose an instance uniformly at random from the local dataset $M_i$. \\
  (b) Set $M_i^{+}=\Bigl\{ (\mathbf{x},y) \in M_i  : y \langle\hat{\mathbf{w}}_i^{(t)},\mathbf{x}\rangle \!<\! 1\Bigr\}$ \\
  (c) Set $\hat{L_i}^{(t)}=  y\mathbf{x}$ \\
  (d) Set $\alpha^{(t)}=\frac{1}{\lambda t}$\\
  (e) Set $\tilde{\mathbf{w}}_i^{t+\frac{1}{2}}=(1-\lambda\alpha^{(t)})\hat{\mathbf{w}}_i^{(t)}+ \alpha^{(t)} \hat{L_i}^{(t)}$ \\
  (f) [Optional] Set  $\tilde{\mathbf{w}}_i^{(t+\frac{1}{2})}=$min$ \Biggl\{1,\frac{\frac{1}{\sqrt{\lambda}}}
{||\tilde{\mathbf{w}}_i^{(t+\frac{1}{2})}||}\Biggr\}  \tilde{\mathbf{w}}_i^{(t+\frac{1}{2})}$ \\
 (g) Set $\hat{\mathbf{w}}_i^{(t+\frac{1}{2})} \gets $PS$(B,\tilde{\mathbf{w}}^{t+\frac{1}{2}})$ \\
 (h) [Optional] Set  $\hat{\mathbf{w}}_i^{(t+1)}=$min$ \Biggl\{1,\frac{\frac{1}{\sqrt{\lambda}}}
{||\hat{\mathbf{w}}_i^{(t+\frac{1}{2})}||}\Biggr\}  \hat{\mathbf{w}}_i^{(t+\frac{1}{2})}$ \; 
  }
\caption{GADGET SVM Algorithm}
\label{gadget}
}
\end{algorithm}

\noindent \textbf{Algorithm Description. }The distributed SVM algorithm (described in Table~\ref{gadget}) takes as input the following parameters: $\lambda$ -- the learning rate, $\mathcal{T}$ - the number of iterations to perform, and $B$ -- a doubly stochastic transition probability matrix. It proceeds as follows: each site $S_i$ builds a linear SVM model on its local data $M_i$ by learning a weight vector $\hat{\mathbf{w}}_i^{(t)}$ at iteration $t$ of the algorithm. The approximate local loss $\hat{L}_i^{(t)}$ corresponding to the current weight vector $\hat{\mathbf{w}}_i^{(t)}$ is estimated. At iteration $t+1$, the local weight vector is updated by taking a step in the direction of the sub-gradient. This intermediate weight vector, $\tilde{\mathbf{w}}_i^{(t+\frac{1}{2})}$, depends on the learning rate $\lambda$ and the approximate loss estimated at iteration $t$. In particular, it is updated by the following sub-gradient update rule:

\begin{equation}
\mathbf{\tilde{w_i}}^{(t+\frac{1}{2})} = (1 - \lambda \alpha^{(t)})\mathbf{\hat{w}_i}^{(t)} + \alpha^{(t)}\hat{L}_{i}^{(t)}
\end{equation}

\noindent Site $S_i$  then gossips the  learnt $\tilde{\mathbf{w}}_i^{(t+\frac{1}{2})}$ with a randomly chosen neighbor using protocol Push-Sum (PS). Protocol Push-Sum takes as input the doubly stochastic $m \times m$ matrix $B$ that stores the transition probability between sites, in addition to the approximate weight vector $\tilde{\mathbf{w}}_i^{(t+\frac{1}{2})}$. On termination of the network wide Push-Sum protocol, the local weight vector at site $i$,   $\hat{\mathbf{w}}_i^{(t+\frac{1}{2})}$ is updated by projecting $\mathbf{\hat{w}}_i^{(t+\frac{1}{2})}$ onto the ball of radius $1/\sqrt{\lambda}$ in order to bound the maximum sub-gradient, in the same spirit as in the Pegasos algorithm (\citet{Singer_07}).

The Push-Sum protocol (\citet{Kempe_03})
deterministically simulates a random walk across $G$ and estimates network sums. 
If an arbitrary stochastic matrix $B = (b_{i,j})$ is created for the network ensuring that if there 
is no edge from $i$ to $j$ in $G$, then $b_{i,j} = 0$, and otherwise $B$ is ergodic and reversible,
then Push-Sum converges to a $\gamma$-relative error solution in $O(\tau_{mix} \log \frac{1}{\gamma})$,
where $\tau_{mix}$ is the mixing speed of the Markov Chain defined by $B$ (\citet{Dutta_17a}). Informally, $\tau_{mix}$
is the the time until $B$ is ``close'' to its steady state distribution. An obvious choice for $B$ 
is to use the random walk on the underlying topology, i.e. $b_{i,j} = \frac{1}{\deg i}$. In general, 
the nodes are not expected to know $\tau_{mix}$, and a simple technique with 
multiplicative overhead for the nodes to calculate a stopping time for Push-Sum (assuming an upper bound on network diameter) has been proposed in work done by \citet{Kempe_03}. 


We should note that the use of an approximate consensus protocol like Push-Sum is necessary because
each node maintains an $\epsilon$-accurate global solution
sum without complete knowledge of the network. 
Providing nodes with this information is problematic because it is liable to change during
operation, and requires additional communication than otherwise necessary.

\subsection{Analysis}
Before analyzing the GADGET algorithm, the notion of strong convexity and
 a sub-differential needs to be introduced. These are essential tools for obtaining bounds on the 
convergence rate of the algorithm in addition to understanding the convergence of the projected sub-gradient method and the network error. We will prove a lemma about the SVM loss function and then apply the 
relative error bounds of Push-Sum to obtain the true convergence rate. 

Informally, a strong convex function is a function who's gradient is always changing, or
equivalently the Hessian is always positive definite. Unfortunately, the SVM objective
function is not differentiable and our analysis must rely on the following, more general, definition of 
strong convexity using sub-differentials. The following 
is the formal definition of strong convexity.
 
\begin{definition}
A function $f: \Re^d \to \Re$ is $\lambda$-strongly convex if $\forall x,y \in \Re^d$ and $\forall g \in \partial f(x)$,\\
$f(y)\ge f(x)+\langle g,y-x \rangle+\frac{\lambda}{2}||x-y||^2.$\\  
\label{def}
\end{definition}

\begin{definition}
\label{subDif}
The \textit{sub-differential} of $f(x)$, denoted 
$\partial f(x)$, is the set of all tangent lines which can be drawn under $f(x)$. Moreover, each 
vector $v \in \partial f(x)$ is called a \textit{sub-gradient} of $f(x)$.
\end{definition}

%

\begin{lem}
\label{lem1}
If $L_g(w)$ is the global average loss of the vector $w$, then we have\\ $ ||L_g(w_1)-L_g(w_2)|| \leq R||w_1-w_2||_{2}.$
\end{lem}

\noindent \emph{Proof: }
The global hinge loss is equal to $\frac{1}{n} \sum_{i=1}^n \max \{0, 1 - y_i \langle x_i, w \rangle \}$; Using triangle inequality we have
$\Norm{L_g(w_1) - L_g(w_2)}_2 \le \frac{1}{n} \sum_{i=1}^n \Norm{ \max \{0, 1 - y_i \langle x_i, w_1\rangle \} - \max \{0, 1 - y_i \langle x_i, w_2\rangle \}}_2.$ Notice that if one of the max functions is zero (assume without loss of generality $1 - y_i\langle x_i, w_2 \rangle \le 0$), the difference of loss equation (for a particular feature) is less or equal to $\Norm{(1 - y_i \langle x_i, w_1\rangle) - (1 - y_i \langle x_i, w_2 \rangle)}_2$. Further observing that if both losses are non-zero, we also arrive at this same equation, the difference in global loss simplifies to 
$\frac{1}{n} \sum_{i=1}^n \Norm{y_i \langle x_i, w_2\rangle - y_i \langle x_i, w_1\rangle}_2 \le \frac{1}{n} \sum_{i=1}^n \Norm{\langle x_i, w_2 - w_1\rangle}_2$.
 Finally, using the Cauchy-Schwartz inequality, we obtain our desired result. 
 \hfill$\square$

\begin{lem}
\label{lem2}
Let $\mathbf{v}_{t,i}$ be the $k \times 1$ vector held by the node i after the $t^{th}$ iteration of Push-Vector, $w_{t,i}$ its weight at that time, and $\mathbf{v}$ the correct vector. Define $M=\sum_{i}|\mathbf{v}_{0,i}|$ to be the vector whose (r,1) entry is the sum of the absolute values of the initial vector $\mathbf{v}_{0,i}$ at all nodes i. Then, for any $\epsilon$, the approximation error is $||\frac{\mathbf{v}_{(t,i)}}{w_{(t,i)}}-\mathbf{v}||_{2} \leq \epsilon||M||_2$, after t=O(${\tau}_{mix}\cdot \log \frac{1}{\epsilon}$)
\end{lem}

\begin{thm}
\label{thm1}
Assuming that $||\mathbf{w}^{(t)}-\hat{\mathbf{w}}_i^{(t)}|| < \epsilon$, we have\\
\begin{align}
\begin{split}
||\mathbf{w}^{(t+\frac{1}{2})}-\hat{\mathbf{w}}_i^{(t+\frac{1}{2})}||\leq  (1-\lambda\alpha^{(t)})(\epsilon+\frac{N\epsilon_1}{\lambda})+\alpha^{(t)}(\frac{R\epsilon}{N}+\epsilon_2)
\end{split}
\end{align}
where $\epsilon_2$ is the relative error in approximating average global loss through Push-Sum of loss on each node and $\epsilon_1$ is the relative error in approximating the weighted sum of the weight (support) vector estimated at each node i.e. $\frac{\sum_{i}n_i\hat{\mathbf{w}}_i^{(t)}}{N}$ 
\end{thm}

\noindent \emph{Proof: } 
\begin{align}
\begin{split}
||\mathbf{w}^{(t+\frac{1}{2})}-\hat{\mathbf{w}}_i^{(t+\frac{1}{2})}||
&=||(1-\lambda\alpha^{(t)})\mathbf{w}^{(t)}+\alpha^{(t)}\frac{L^{(t)}}{N}-(1-\lambda\alpha^{(t)})PS(n_i\hat{\mathbf{w}}_i^{(t)},B)-\alpha^{(t)}PS(\hat{L}_i^{(t)},B)||\\
&\leq (1-\lambda\alpha^{(t)})(||\mathbf{w}^{(t)}-\frac{\sum_{i}n_i\hat{\mathbf{w}}_i^{(t)}}{N}||+||\frac{\sum_{i}n_i\hat{\mathbf{w}}_i^{(t)}}{N}-PS(n_i\hat{\mathbf{w}}_i^{(t)},B)||)\\
&+\alpha^{(t)}(||\frac{L^{(t)}}{N}-\frac{\sum_i\hat{L}_i^{(t)}}{N}||+||\frac{\sum_i\hat{L}_i^{(t)}}{N}-PS(\hat{L}_i^{(t)},B)||)\\
& \leq (1-\lambda\alpha^{(t)})(\epsilon+\frac{N\epsilon_1}{\lambda})+\alpha^{(t)}(\frac{R\epsilon}{N}+\epsilon_2)
\end{split}
\end{align}
where we have used Lemma 2, Lemma 3, and the following:

$||\mathbf{w}^{(t)}-\frac{\sum_{i}n_i\hat{\mathbf{w}}_i^{(t)}}{N}|| = ||\sum_i { \frac{n_i\mathbf{w}^{(t)}} {N} } -  \frac{\sum_{i}n_i\hat{\mathbf{w}}_i^{(t)}}{N}    || \leq\sum_i \frac{n_i\epsilon}{N}=\epsilon$
$$||\frac{\sum_{i}n_i\hat{\mathbf{w}}_i^{(t)}}{N}-PS(n_i\hat{\mathbf{w}}_i^{(t)},B)||\\
\leq \epsilon_1\frac{1}{N^2}\sum_{k=1}^{K}\bigg(\sum_{i=1}^{N}|n_i\hat{\mathbf{w}}_{i}^{(t+\frac{1}{2})}(k)|\bigg)^{2}$$\\
$$\leq\frac{\sum_i n_i^2}{N^2}\sum_{i=1}^{N}\sum_{k=1}^{K}
(\hat{\mathbf{w}}_{i}^{(t+\frac{1}{2})}(k))^2\leq\frac{\sum_in_i^2}{N^2}\frac{N}{\lambda}\leq\frac{N}{\lambda}$$\\
$$||\frac{\sum_i\hat{L}_i^{(t)}}{N}-PS(\hat{L}_i^{(t)},B)||\leq \frac{\epsilon_2}{N}\sqrt{\sum_{i=1}^{N}(\hat{L}_i^{(t)})^2}\leq\epsilon_2$$
\hfill$\square$

\begin{lem}
Let $\mathbf{w}^{(t)}$ and $\mathbf{u}$ be two weight vectors, and lets $\nabla_t$ denotes the sub-gradient at $\mathbf{w}^{(t)}$, then 
$$\langle\mathbf{w}^{(t)}-\mathbf{u}\rangle\leq\frac{||\mathbf{w}^{(t)}-\mathbf{u}||^2-||\mathbf{w}^{(t+1)}-\mathbf{u}||^2}{2\alpha^{(t)}}+\frac{\alpha^{(t)}}{2}c^2$$
\label{imp}
\end{lem}

\noindent \emph{Proof: } The proof follows from Lemma 1. in \citet{Shwartz_2007}.

\begin{thm}
Assume that the conditions in Theorem~\ref{thm1} hold. Then it can be shown that $f(\displaystyle\frac{\bar{\mathbf{w}}_i}{T})-f(\mathbf{w}^{\star})\leq\frac{2c}{\sqrt{\lambda}}+\frac{c^2 log(T)}{2T\lambda}+\frac{2}{\sqrt{\lambda}}\bigg(\frac{\gamma R}{\sqrt{\lambda}}+\gamma R\bigg)$
\end{thm}

\noindent \emph{Proof: } The proof is presented in the Appendix~\ref{appendA}.

\section{Experimental Results}
\label{empirical}

\subsection{Aims}
\label{sec:exptaims}
Our objective is to investigate empirically the utility of the GADGET SVM
algorithm we have described. We intend to examine if there is empirical support for the conjecture
that the performance of the $Distributed$ model is better than that of the $Centralized$\footnote{Our centralized model is based on the execution of the Pegasos algorithm (\cite{Shwartz_2007}) on the entire dataset on a single node.} model.
We are assuming that the performance of a
model-construction method is given by the pair $(A,T)$
where $A$ is an unbiased estimate of the predictive accuracy of
the classifier, and $T$ is an unbiased estimate of the
time taken to construct a model. In all cases, 
the time taken to construct a model does not include the time to read the dataset into local memory. Comparison of
pairs $(A_1,T_1)$ and $(A_2,T_2)$ will simply be lexicographic comparisons. 

\begin{figure}[!th]
\centering
\subfigure
  \centering
  \includegraphics[width=0.98\linewidth, height=0.32\textheight]{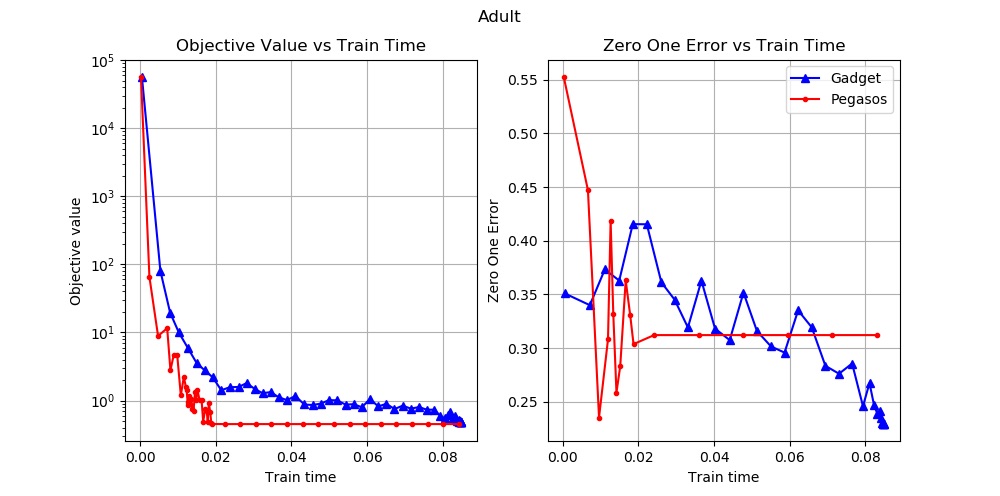}
  \label{fig:adult}
 \subfigure
  \centering
  \includegraphics[width=0.98\linewidth,height=0.32\textheight]{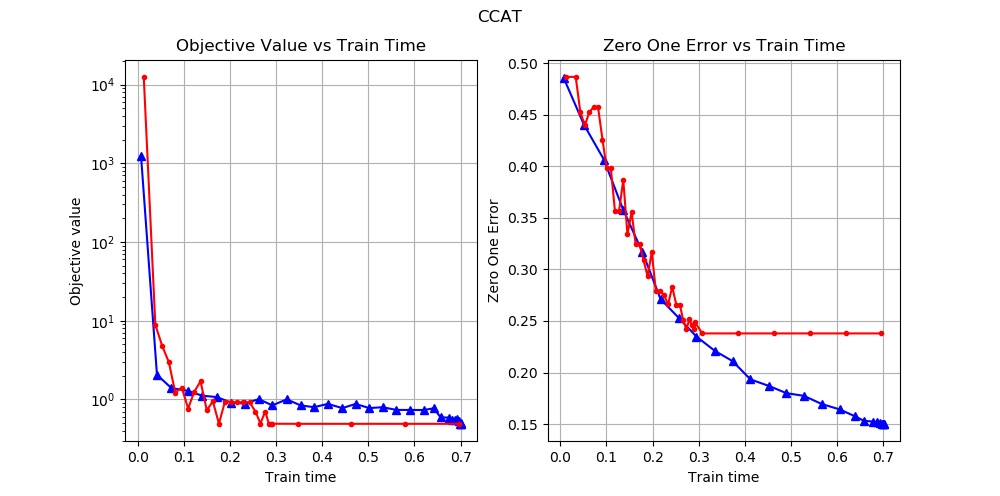}
  \label{fig:ccat}

\label{sub:AllOne} 
\end{figure}

\subsection{Materials}
\label{data}
\noindent \textbf{Data}
The datasets used for our experiments are described below. All of them were obtained from the following website: \url{http://leon.bottou.org/papers/bordes-ertekin-weston-bottou-2005\#data\_sets} except CCAT\footnote{\url{http://www.ai.mit.edu/projects/jmlr/papers/volume5/lewis04a/lyrl2004_rcv1v2_README.htm}} , USPS, and Webspam datasets.
\begin{itemize}

\item \emph{CCAT: }This is a text categorization task taken from the Reuters RCV1 collection. The training set consists of $781265$ examples, and the test set has $23148$ examples. The original dataset has several topics associated with each example -- this was transformed into a binary classification task by assigning a positive label to any example which had CCAT as one of its labels, and a negative label to all other examples.

\item \emph{Reuters: } This task makes use of the Reuters-21578 dataset, a popular collection for text-categorization research. A binary classification task is designed by splitting the data with respect to money-fx labels versus all others. The training set has $7770$ examples, and the test set $3299$ examples.

\item \emph{Adult: } The Adult dataset has been extracted from the census bureau database found at \texttt{\url{http://www.census.gov/ftp/pub/DES/www/welcome.html}}. The task is to predict whether a person makes above $\$50000$ a year using $14$ attributes that include race, sex, occupation and others. This dataset comprises of $32562$ training examples and $16282$ test examples. 

\item \emph{MNIST: } The MNIST handwritten digits dataset consists of $60000$ training and $10000$ test examples. It consists of images of size $28 \times 28$, and the task is to predict the digit represented by the image, which can lie between $0$ and $9$. To design a binary classification task, we choose to predict whether or not the digit $0$ is present in the image. This results in training examples with a total of $784$ attributes.

\item \emph{USPS: } This dataset\footnote{http://leon.bottou.org/papers/bordes-ertekin-weston-bottou-2005\#data\_sets} is obtained by scanning handwritten digits from envelopes by the U.S. Postal Service. Similar to the MNIST dataset, the label ``0" versus the rest is used for designing a binary classification task.

\item \emph{Webspam: }The webspam or the Webb Spam corpus (\cite{Wang_12a}) consists of webpages that belong to the spam and non-spam categories. The unigram version of the dataset, which contains 350000 examples, and 254 features is used for our experiments. We randomly split this data set into train and test partitions with a 2:1 ratio. The dataset contains 234500 train examples 115500 test examples. 

\end{itemize}

\begin{figure}[!h]
\centering
 \subfigure
  \centering
  \includegraphics[width=0.98\linewidth,height=0.32\textheight]{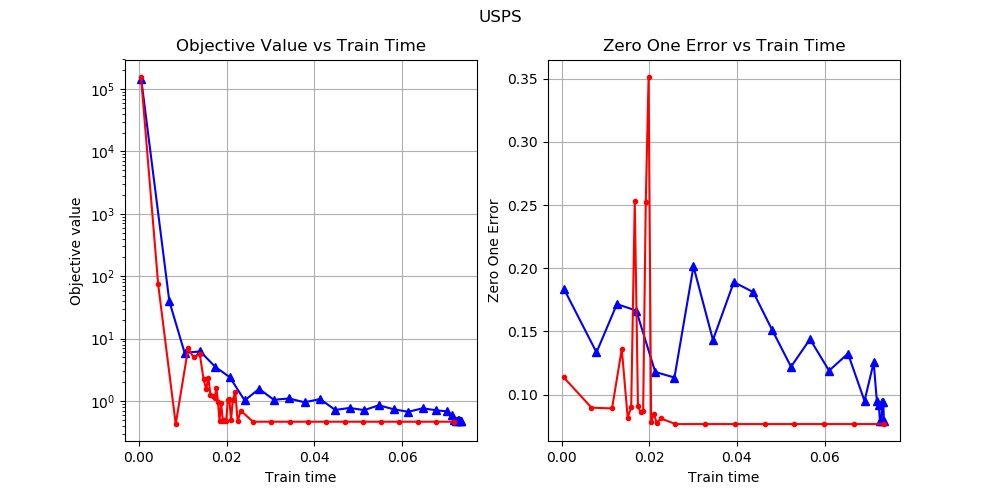}
  \label{fig:usps}
 \subfigure
  \centering
  \includegraphics[width=0.98\linewidth,height=0.32\textheight]{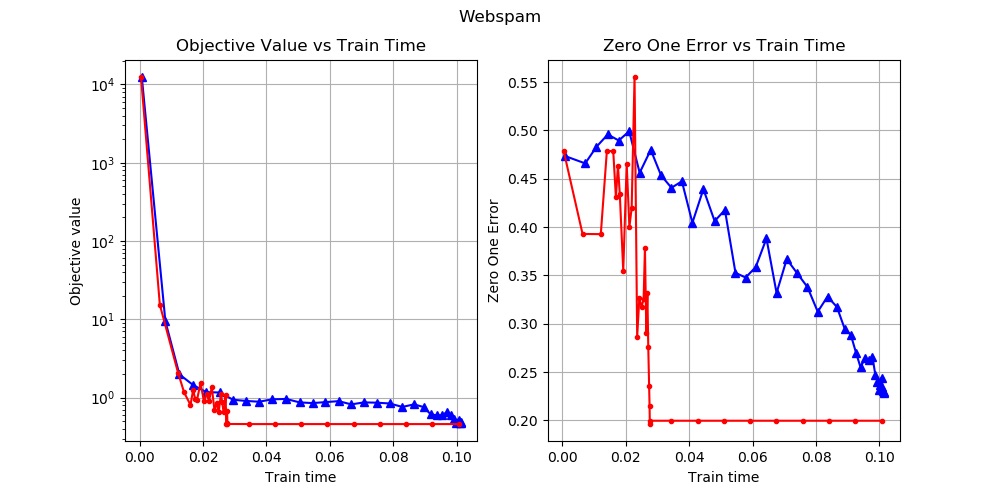}
  \label{fig:webspam}
 \label{sub:AllTwo} 
\end{figure}

The datasets and their properties are summarized in the Table~\ref{fig:datatable}. 

\begin{table}[h]

\begin{center}

\begin{tabular}{|c|c|c|c|c|c|c|c|} \hline
Dataset & Training Size & Testing Size & Features & Sparsity & $\lambda$  \\ \hline
Adult  &            32561 &               16281 &           123 &                  NA       & $3.07 \times 10^{-5}$   \\ \hline 
CCAT  &            781265 &              23149 &          47236 &               $0.16\%$ & $10^{-4}$   \\ \hline
MNIST  &          60000  &              10000 &           784 &               NA        & $1.67 \times 10^{-5}$   \\ \hline 
Reuters  &          7770 &                3299 &            8315 &                 NA        & $1.29 \times 10^{-4}$   \\ \hline 
USPS  &          7329  &              1969 &           256 &               NA         & $1.36 \times 10^{-4}$    \\ \hline 
Webspam &  234500 & 115500 & 254 & NA &  $10^{-5}$\\ \hline
\end{tabular}
\end{center}
\caption{Summary of the characteristics of the data used for empirical analysis.}
\label{fig:datatable}
\end{table}%

\subsection{Algorithms and Machines}
The GADGET algorithm has been implemented on Peersim\footnote{\url{http://peersim.sourceforge.net/}}, a peer-to-peer network(P2P) simulator. This software\footnote{The code is available from \texttt{https://github.com/nitinnat/GADGET}.} allows simulation of the P2P network by initializing nodes and the communication protocols to be used by them. GADGET implements a cycle driven protocol that has periodic activity in approximately regular time intervals. Nodes are able to communicate with others using the Push-Sum protocol (described in Section~\ref{gossip}). The experiments are performed on a single node, on a DELL E7-4830 architecture, equipped with 32 x 2.13GHz Intel Xeon CPU E7-4830 Processor Cores; instruction and data cache sizes of 24576 KB; a main memory size of 512 GB and 128 GB RAM. The computational resources were provided by The Center for Computational Research at the State University of New York (SUNY) at Buffalo\footnote{\url{https://www.buffalo.edu/ccr.html}}.  

\begin{figure}[!h]
\centering
 \subfigure
  \centering
  \includegraphics[width=0.98\linewidth,height=0.32\textheight]{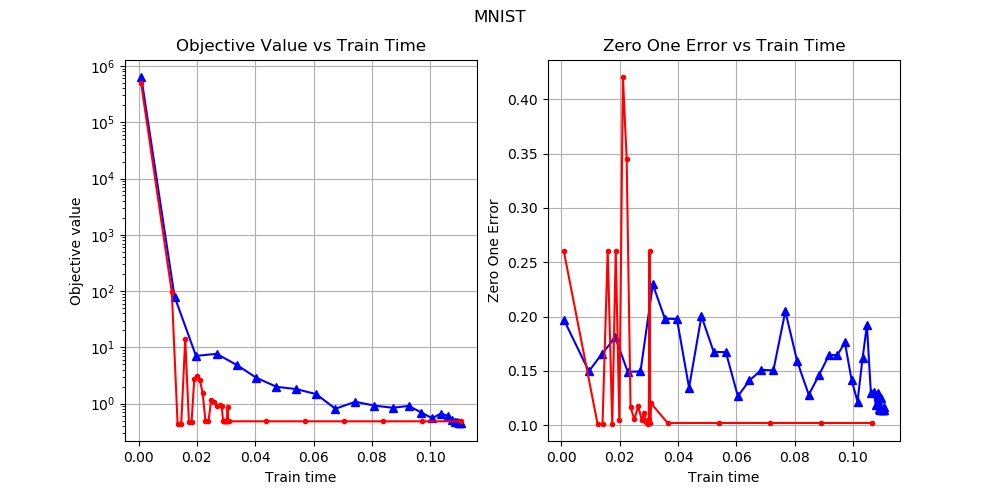}
  \label{fig:mnist}
  \subfigure
  \centering
  \includegraphics[width=0.98\linewidth, height=0.32\textheight]{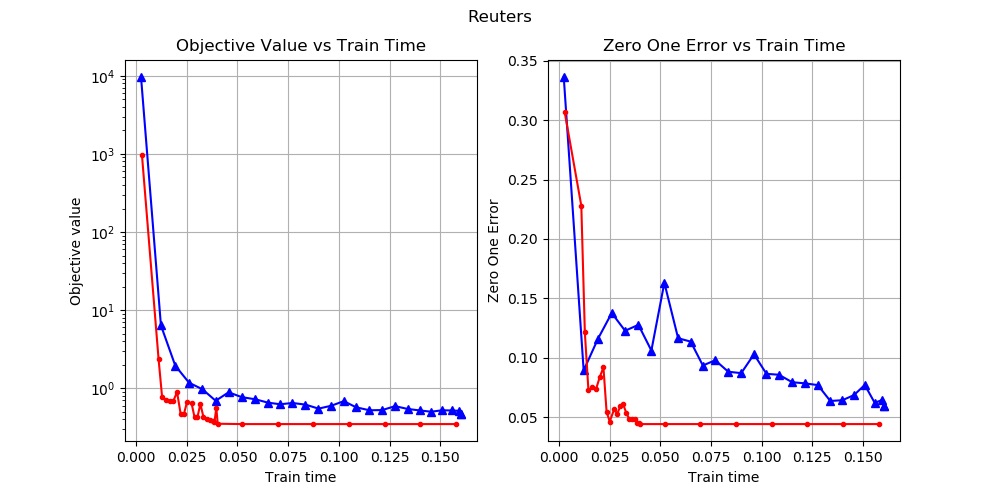}
  \label{fig:reuters}
 \label{sub:AllThree} 
\end{figure}


\begin{table}[!h]
\begin{center}
\begin{tabular}{|c|cc|cc|} \hline
Dataset & \multicolumn{2}{l|}{GADGET}& \multicolumn{2}{l|}{Pegasos (Centralized)} \\ \hline
& Time (secs) & (Acc. $\%$) & Time (secs) & (Acc.$\%$)  \\ \hline
Adult & 0.08 ($\pm 0.01$) & 77.04 ($\pm 0.03$) & 0.02 ($\pm 0.002$) & 68.79 ($\pm 0.18$) \\ \hline
CCAT & 0.70 ($\pm 0.101$) & 84.99 ($\pm 0.016$) & 0.29 ($\pm 0.031$)  & 76.21 ($\pm 0.04$) \\ \hline
MNIST & 0.11 ($\pm 0.037$) & 88.57 ($\pm 0.04$) & 0.03 ($\pm 0.004$) & 89.81 ($\pm 0.01$)\\ \hline
Reuters & 0.16 ($\pm 0.01$) & 94.04 ($\pm 0.07$) & 0.04 ($\pm 0.006$) & 95.59 ($\pm 0.01$)\\ \hline
USPS & 0.07 ($\pm 0.013$) & 92.12 ($\pm 0.03$) &0.023 ($\pm 0.003$) & 92.33 ($\pm 0.02$) \\ \hline
Webspam & 0.10 ($\pm 0.02$) & 77.49 ($\pm 0.05$) & 0.03 ($\pm 0.003$) & 80.04 ($\pm 0.04$)  \\ \hline
\end{tabular}
\end{center}
\caption{Comparison of the performance of GADGET SVM and centralized Pegasos on five data sets using the following metrics: accuracy of classification, time taken for the distributed algorithm versus the centralized, and percentage of speed-up in execution time. The accuracy and time reported for GADGET is the mean over all the nodes in the network over five different trials. The standard deviations reported are estimated by $\sqrt{Var(Nodes) + Var(Trials)}$ where $Nodes$ represent the number of nodes in the network and $Trials$ refer to the number of trials of each experiment. Epsilon at convergence  of GADGET is $\epsilon$ = 0.000862047, 0.00087048, 0.000671476, 0.000961603, 0.00092748, 0.00081329 for the Adult, CCAT, MNIST, Reuters, USPS, and Webspam datasets respectively.}
\label{fig:tbPerf}
\end{table}%

\subsection{Method}
The following method was executed on each of the datasets mentioned above (Section~\ref{data}):
\begin{enumerate}
\item A network of $k$ nodes is setup using the Peersim simulator.
\item The train and test sets are split into $k$ different files, containing approximately equal number of instances and distributed amongst the nodes.
\item The GADGET algorithm (Section~\ref{prelim}) is executed on each node independently until the local weight vectors converge i.e. they do not change more than an user-defined parameter $\epsilon$. The local models are then used to determine the primal objective and the test error on the corresponding test sets. The results are averaged over all the nodes in the network. This entire process is executed several times, and an average value of primal objective and test error is obtained.
\item In addition, the Pegasos algorithm is executed on the entire dataset -- this serves as a baseline against which the performance of GADGET is evaluated. Figures~\ref{sub:AllOne},~\ref{sub:AllTwo}, and~\ref{sub:AllThree} present the plots of objective value and zero-one error versus train time of the distributed models.
\end{enumerate}
 

The following details are relevant:
\begin{enumerate}
\item  $k=10$ in the experiments reported in this paper.
\item The accuracy and time reported in Table~\ref{fig:tbPerf} are averages (and corresponding standard deviations) over all the nodes in the network over five trials. 
\item The user-defined $\epsilon$ parameter is set to be $0.001$.
\item The $\lambda$ values for all the datasets were chosen to be identical to benchmark tests performed in \cite{Shwartz_2007}, and can be found in Table~\ref{fig:datatable}.
\end{enumerate}



\subsection{Results}

\subsubsection{Comparison between distributed and centralized algorithms}
Table~\ref{fig:tbPerf} summarizes the average accuracy obtained on the test data distributed among nodes. It is observed that the performance (as measured by accuracy of classification) of the GADGET SVM is comparable to the centralized algorithm for all practical purposes. The training time of the centralized model often outperforms the distributed one when data loading time is not included -- this is because of at least two reasons: (a) The centralized algorithm simply estimates the approximate gradient on a randomly chosen sample from the entire dataset; GADGET has to repeat this step independently at all the nodes in the system -- we report average time taken across the nodes along with standard deviations. (b) GADGET communicates the weight vector amongst nodes thereby incurring a larger communication cost.


\begin{table}[!h]
\begin{center}
\begin{minipage}[b]{1\linewidth}
\small
{
\begin{tabular}{|c|ll|ll|ll|} \hline
Dataset & \multicolumn{2}{c|}{GADGET} & \multicolumn{2}{c|}{$SVM^{Perf}$} & \multicolumn{2}{c|}{SVM-SGD} \\ \hline
& Time (secs) & (Acc. $\%$) & Time (secs) & (Acc.$\%$)  & Time (secs) & (Acc. $\%$)\\ \hline
Adult & 0.085 ($\pm0.01$) & 77.04 ($\pm0.03$) & 0.059($\pm0.01$) & 76.38 ($\pm1.00$)  & 0.025 ($\pm0.006$)& 84.14($\pm0.88$)  \\ \hline
CCAT & 0.70 ($\pm 0.10$) & 84.99 ($\pm 0.02$) & 4.33($\pm0.49$) & 53.41($\pm1.062$) & 6.64($\pm0.16$) & 92.82($\pm0.52$)\\ \hline
MNIST & 0.11 ($\pm 0.04$) & 88.57 ($\pm 0.04$) & 0.81($\pm 1.81$) & 89.91($\pm0.83$)  & 0.83($\pm0.03$)& 82.52($\pm2.69$)\\ \hline
Reuters & 0.16 ($\pm 0.01$) & 94.04 ($\pm 0.07$) & 0.05($\pm0.01$) &94.57($\pm1.22$) & 0.04($\pm0.00$) & 94.35($\pm0.88$) \\ \hline
USPS & 0.072 ($\pm 0.013$) & 92.12 ($\pm 0.03$) & 18.64($\pm 22.89$) & 91.62($\pm1.90$)  & 0.15($\pm0.008$) & 91.63($\pm1.96$)\\ \hline
Webspam & 0.10 ($\pm 0.02$) & 77.49 ($\pm 0.05$) & 1.38($\pm 0.11$) & 60.72 ($\pm 0.42$) & 0 2.52 ($\pm 0.00$) & 89.94 ($\pm 2.19$) \\ \hline
\end{tabular}
}
\end{minipage}
\end{center}
\caption{Comparison of the performance of GADGET SVM, $SVM^{Perf}$ and SVM-SGD. All algorithms are executed individually on each node of the network.}
\label{tbComp}
\end{table}%

\subsubsection{Comparison between GADGET SVM and state-of-the-art online SVM algorithms} We compare the performance of GADGET SVM to two state-of-the-art online SVM algorithms -  $SVM^{Perf}$ (\cite{Joachims_09a,Joachims_06}) and SVM SGD(\cite{bottou-2010, bottou-98x}). We chose these algorithms for the following reasons: (a) Since GADGET SVM is designed for primal SVM formulations, we preferred to compare it against state-of-the-art primal solvers with $L_1$ regularization. (b) Given that GADGET SVM uses a stochastic gradient descent method, its performance is compared against other SGD solvers. We studied separately what would have happened if $SVM^{Perf}$ and SVM SGD received random instances which the algorithm classified -- to enable this process, each node executed the algorithm and reported a test performance. This in effect means that the two online algorithms will execute in a ``distributed" fashion, albeit, without communication amongst the nodes. To the best of our knowledge, there are no known theoretical guarantees on global convergence in these settings for either $SVM^{Perf}$ or SVM SGD. Thus this setting is somewhat different from the gossip-based consensus setting in which GADGET SVM is executed. However, this appears to be the best choice in terms of comparison of the proposed algorithm against other distributed algorithms\footnote{We have explored options of comparing GADGET with another consensus-based SVM(\cite{Forero_10}) algorithm; however, the code is not available and hence not replicable (personal communication with authors)}. Table~\ref{tbComp} presents the results. With regard to classification accuracy, GADGET SVM has comparable or better performance than both $SVM^{Perf}$ and SVM-SGD. In three of the six datasets, SVM SGD has a better accuracy of prediction -- however, multiple trials revealed a large standard deviation on this result. SVM SGD also proved to be faster than GADGET -- this is expected since the gossip protocol uses additional information from nodes in the network which are then used to update local models. There is a higher communication overhead that the protocol needs to deal with in the distributed setting.
 $SVM^{Perf}$ was noted to take substantially longer than both GADGET SVM and SVM SGD. These results suggest that the GADGET SVM algorithm can provide comparable accuracy to state-of-the-art solvers within reasonable time to construct the model. This is very useful in distributed and resource constrained environments where centralization of data may not be an option and distributed algorithms are the norm. 

\section{Conclusions}
\label{conc}
We presented a distributed algorithm, GADGET SVM, for approximately minimizing the objective function of a linear SVM using the primal formulation. The algorithm  uses a gossip-based protocol to communicate amongst distributed nodes. We derived theoretical bounds to show that the algorithm is guaranteed to converge and presented empirical results on seven publicly available data sets. Our results indicate that the accuracy of the distributed algorithm is comparable to state-of-the-art centralized counterparts (such as Pegasos) and online variants including SVM SGD and SVM Perf. 

There are several directions for future work including studying the effect of other optimization algorithms (such as mini-batch variants of stochastic gradient descent, coordinate descent) on performance of distributed algorithm, extension to multi-class variants of SVMs, resilience to node failures, impact of the underlying network structure on the convergence of the algorithm and development of distributed gossip-based algorithms for non-linear SVMs. We are hope to address these in sequel.

\ACKNOWLEDGMENT{%
This research is supported by a summer research grant from the School of Management, University at Buffalo. H.D. would like to thank (late) Dr. David Waltz for the inspiration and advice to develop this project. Chase Hensel, Deepak Nayak, Suraj Kesari, and Raghuram Nagireddy helped with software development and interesting discussions during various phases of the project.

}

%
%
%


\bibliographystyle{informs2014} 
\bibliography{consensus} 

\newpage
 \begin{APPENDIX}{}
\section{Appendix A}
\label{appendA}

\begin{thm}
Assume that the conditions in Theorem~\ref{thm1} hold. Then it can be shown that 
\begin{equation}
\nonumber
f(\displaystyle\frac{\bar{\mathbf{w}}_i}{T})-f(\mathbf{w}^{\star})\leq\frac{2c}{\sqrt{\lambda}}+\frac{c^2 log(T)}{2T\lambda}+\frac{2}{\sqrt{\lambda}}\bigg(\frac{\gamma R}{\sqrt{\lambda}}+\gamma R\bigg)
\end{equation}
\end{thm}

\noindent \emph{Proof: }
From Algorithm 2, we have,
\begin{align}
\begin{split}
\tilde{\mathbf{w}}_i^{(t+\frac{1}{2})}=(1-\lambda\alpha^{(t)})n_i\hat{\mathbf{w}}_i^{(t)}+ \alpha^{(t)} \hat{L_i}^{(t)}\\
\label{a}
\end{split}
\end{align}

\noindent Summing over iterations $ t=1,2,\ldots,T$ we define,
\begin{align}
\begin{split}
\displaystyle\bar{\mathbf{w}}^{(t+\frac{1}{2})}\vcentcolon=\sum_{i=1}^{m}\frac{\hat{\mathbf{w}}_i^{(t+\frac{1}{2})}}{N}
&=\tilde{\bar{\mathbf{w}}}^{(t)}-\alpha^{(t)}\bigg(\lambda\tilde{\bar{\mathbf{w}}}^{(t)}-\hat{L}_g^{(t)}\bigg)\\
\label{b}
\end{split}
\end{align}
where $\tilde{\bar{\mathbf{w}}}^{(t)}=\displaystyle\frac{\sum_{i=1}^{m}n_i\hat{\mathbf{w}}_i^{(t)}}{N}$, $\hat{L}_g^{(t)}=\frac{1}{N}\sum_{i=1}^{m}\hat{L}_i^{(t)}$.
Note that $||\mathbf{w}^{\star}||\leq\frac{1}{\sqrt{\lambda}}
$ and $||\hat{\mathbf{w}}_i^{(t)}||\leq\frac{1}{\sqrt{\lambda}}$ from the algorithm. Also, we have

\begin{align}
\begin{split}
||\tilde{\bar{\mathbf{w}}}^{(t)}||=||\frac{\sum_{i=1}^{m}n_i\hat{\mathbf{w}}_i^{(t)}}{N}||\leq\sum_{i=1}^{m}\frac{n_i}{N}||\hat{\mathbf{w}}_i^{(t)}||\leq\sum_{i=1}^{m}\frac{n_i}{\sqrt{\lambda}N}=\frac{1}{\sqrt{\lambda}}\\
\end{split}
\end{align}

Assuming that $c$ is the maximum sub-gradient, and using the fact that increment of a continuous function over an interval is always less than its maximum subgradient times the length of the interval, we have\\
\begin{align}
\begin{split}
f(\hat{\mathbf{w}}_i^{(t)})-f(\tilde{\bar{\mathbf{w}}}^{(t)})\leq c||\hat{\mathbf{w}}_i^{(t)}-\tilde{\bar{\mathbf{w}}}^{(t)}||_2
\label{c}
\end{split}
\end{align}

where $f$ represents Eq.(1). Using $\lambda$-strong convexity of $f$ from definition \ref{def} with $x=\tilde{\bar{\mathbf{w}}}^{(t)}, y=\mathbf{w}^{\star}$, and $g=\lambda\tilde{\bar{\mathbf{w}}}^{(t)}-L_g(\tilde{\bar{\mathbf{w}}}^{(t)})$ we have

\begin{align}
\begin{split}
f(\tilde{\bar{\mathbf{w}}}^{(t)})-f(\mathbf{w}^{\star})+\frac{\lambda}{2}||\tilde{\bar{\mathbf{w}}}^{(t)}-\mathbf{w}^{\star}||^2\leq\langle\lambda\tilde{\bar{\mathbf{w}}}^{(t)}-L_g(\tilde{\bar{\mathbf{w}}}^{(t)}),\tilde{\bar{\mathbf{w}}}^{(t)}-\mathbf{w}^{\star}\rangle
\label{d}
\end{split}
\end{align}

Summing over t in Eq. \ref{c} and subtracting $Tf(\mathbf{w}^{\star})$ on both sides we have
\begin{align}
\begin{split}
\sum_{t=1}^{T}f(\hat{\mathbf{w}}_i^{(t)})-Tf(\mathbf{w}^{\star})&\leq c\sum_{t=1}^{T}||\hat{\mathbf{w}}_i^{(t)}-\tilde{\bar{\mathbf{w}}}^{(t)}||+\sum_{t=1}^{T}f(\tilde{\bar{\mathbf{w}}}^{(t)})-Tf(\mathbf{w}^{\star})\\
\label{h}
\end{split}
\end{align}

Using Eq.\ref{d} to replace $f(\tilde{\bar{\mathbf{w}}}^{(t)})-f(\mathbf{w}^{\star})$ in Eq.\ref{h} gives

\begin{align}
\begin{split}
\sum_{t=1}^{T}f(\hat{\mathbf{w}}_i^{(t)})-Tf(\mathbf{w}^{\star})&\leq c\sum_{t=1}^{T}||\hat{\mathbf{w}}_i^{(t)}-\tilde{\bar{\mathbf{w}}}^{(t)}||+\sum_{t=1}^{T}\langle\lambda\tilde{\bar{\mathbf{w}}}^{(t)}-L_g(\tilde{\bar{\mathbf{w}}}^{(t)}),\tilde{\bar{\mathbf{w}}}^{(t)}-\mathbf{w}^{\star}\rangle\\
&-\sum_{t=1}^{T}\frac{\lambda}{2}||\tilde{\bar{\mathbf{w}}}^{(t)}-\mathbf{w}^{\star}||^2\\
\label{f}
\end{split}
\end{align}

Now using Lemma \ref{imp} with $\mathbf{w}^{(t)}=\tilde{\bar{\mathbf{w}}}^{(t)}, \mathbf{u}=\mathbf{w}^{\star}, \nabla_t=\lambda\tilde{\bar{\mathbf{w}}}^{(t)}-\hat{L}_g^{(t)}$, we have
\begin{align}
\begin{split}
\langle\lambda\tilde{\bar{\mathbf{w}}}^{(t)}-\hat{L}_g^{(t)},\tilde{\bar{\mathbf{w}}}^{(t)}-\mathbf{w}^{\star}\rangle & \leq	
\frac{||\tilde{\bar{\mathbf{w}}}^{(t)}-\mathbf{w}^{\star}||^2-||\tilde{\bar{\mathbf{w}}}^{(t+1)}-\mathbf{w}^{\star}||^2}{2\alpha^{(t)}}+\frac{\alpha^{(t)}}{2}c^2\\
\label{12}
\end{split}
\end{align}

Now, using the fact that $\langle a-b,c \rangle = \langle  (a-d)-(b-d),c \rangle=\langle a-d,c \rangle - \langle b-d,c \rangle$, we have

\begin{align}
\begin{split}
\langle\lambda\tilde{\bar{\mathbf{w}}}^{(t)}-L_g(\tilde{\bar{\mathbf{w}}}^{(t)}),\tilde{\bar{\mathbf{w}}}^{(t)}-\mathbf{w}^{\star}\rangle &= \langle\lambda\tilde{\bar{\mathbf{w}}}^{(t)}-\hat{L}_g^{(t)},\tilde{\bar{\mathbf{w}}}^{(t)}-\mathbf{w}^{\star}\rangle+\langle\hat{L}_g^{(t)}-L_g(\tilde{\bar{\mathbf{w}}}^{(t)}),\tilde{\bar{\mathbf{w}}}^{(t)}-\mathbf{w}^{\star}\rangle\\
\label{rand}
\end{split}
\end{align}

Now, $||\tilde{\bar{\mathbf{w}}}^{(t)}-\mathbf{w}^{\star}||\leq||\tilde{\bar{\mathbf{w}}}^{(t)}||+||\mathbf{w}^{\star}||\leq\frac{2}{\sqrt{\lambda}}$. Notice that $||L_g(\tilde{\bar{\mathbf{w}}}^{(t)})- \hat{L}_g^{(t)}||=||L_g(\tilde{\bar{\mathbf{w}}}^{(t)})-\frac{1}{N}\sum_{i=1}^{m}\hat{L}_i^{(t)}||$. Because each global oss approximation is within a $\gamma$-radius ball of the global loss of some w vector which in-turn is within a $\gamma$-radius ball of $\tilde{\bar{\mathbf{w}}}^{(t)}$, this in turn is less than or equal to $||L_g(\tilde{\bar{\mathbf{w}}}^{(t)})- (1-\gamma) L_g((1+\gamma)\tilde{\bar{\mathbf{w}}}^{(t)})||$. Now, using lemma 2 we have the last term $\leq ||L_g(\tilde{\bar{\mathbf{w}}}^{(t)})- L_g((1+\gamma)\tilde{\bar{\mathbf{w}}}^{(t)})|| + \gamma R$. Again applying lemma 2 to the first term, we get it $\leq \frac{\gamma R}{\sqrt{\lambda}}+\gamma R$. Using this bound and Cauchy-Schwartz inequality, we have $\langle\hat{L}_g^{(t)}-L_g(\tilde{\bar{\mathbf{w}}}^{(t)}),\tilde{\bar{\mathbf{w}}}^{(t)}-\mathbf{w}^{\star}\rangle \leq ||\tilde{\bar{\mathbf{w}}}^{(t)}-\mathbf{w}^{\star}||||L_g(\tilde{\bar{\mathbf{w}}}^{(t)})- \hat{L}_g^{(t)}|| \leq \frac{2}{\sqrt{\lambda}} (\frac{\gamma R}{\sqrt{\lambda}}+\gamma R)$. 
Using this and replacing the first term of Eq. \ref{rand} with bound in Eq. \ref{12}, we have

\begin{align}
\begin{split}
\langle\lambda\tilde{\bar{\mathbf{w}}}^{(t)}-L_g(\tilde{\bar{\mathbf{w}}}^{(t)}),\tilde{\bar{\mathbf{w}}}^{(t)}-\mathbf{w}^{\star}\rangle & \leq \frac{||\tilde{\bar{\mathbf{w}}}^{(t)}-\mathbf{w}^{\star}||^2-||\tilde{\bar{\mathbf{w}}}^{(t+1)}-\mathbf{w}^{\star}||^2}{2\alpha^{(t)}}+\frac{\alpha^{(t)}}{2}c^2 \\+
& \frac{2}{\sqrt{\lambda}} (\frac{\gamma R}{\sqrt{\lambda}}+\gamma R)\\
\end{split}
\end{align}

Using this bound in Eq. \ref{f}, and expanding the sum over t, we get\\
\begin{align}
\begin{split}
\sum_{t=1}^{T}f(\hat{\mathbf{w}}_i^{(t)})-Tf(\mathbf{w}^{\star}) \leq
 c\sum_{t=1}^{T}||\hat{\mathbf{w}}_i^{(t)}-\tilde{\bar{\mathbf{w}}}^{(t)}|| + 
(\frac{1}{2 \alpha^{(1)}}-\frac{\lambda}{2} )||\tilde{\bar{\mathbf{w}}}^{(1)}-\mathbf{w}^{\star}||^2 \\
-\bigg (\frac{1}{2\alpha^{(T)}}\bigg )||\tilde{\bar{\mathbf{w}}}^{(T+1)}-\mathbf{w}^{\star}||^2 +
 \frac{c^2}{2}\sum_{t=1}^{T}\alpha^{(t)}\\
+\sum_{t=2}^{T}\bigg(\frac{1}{2\alpha^{(t)}}-\frac{1}{2\alpha^{(t-1)}}-\frac{\lambda}{2}\bigg) 
||\tilde{\bar{\mathbf{w}}}^{(t)}-\mathbf{w}^{\star}||^2 + \frac{2T}{\sqrt{\lambda}} (\frac{\gamma R}{\sqrt{\lambda}}+\gamma R)\\
\label{big}
\end{split}
\end{align}

Substituting in for $\alpha^{(t)} = \frac{1}{\lambda t}$ , we find that the second term and the fifth term is zero. Fourth term is bounded by $\frac{c^2 log(T)}{2\lambda}$. We ignore the third term. Using the fact that $||\hat{\mathbf{w}}_i^{(t)}||$, $||\mathbf{w}^{\star}||$, and $||\tilde{\bar{\mathbf{w}}}^{(t)}||$ are $\leq \frac{1}{\sqrt{\lambda}}$, the first term is bounded by $\frac{2cT}{\sqrt{\lambda}}$. Thus simplifying the Eq. \ref{big} and diving by $T$, we get\\
$$\frac{1}{T}\sum_{t=1}^{T}f(\hat{\mathbf{w}}_i^{(t)})-f(\mathbf{w}^{\star}) \leq\frac{2c}{\sqrt{\lambda}}+\frac{c^2 log(T)}{2T\lambda}+\frac{2}{\sqrt{\lambda}}\bigg(\frac{\gamma R}{\sqrt{\lambda}}+\gamma R\bigg)$$

Using the property of f being a convex function, we have $\frac{1}{T}\sum_{t=1}^{T}f(\hat{\mathbf{w}}_i^{(t)})\leq f(\frac{1}{T}\sum_{t=1}^{T}\hat{\mathbf{w}}_i^{(t)})$. Since $\frac{\bar{\mathbf{w}}_i}{T}=\frac{1}{T}\sum_{t=1}^{T}\hat{\mathbf{w}}_i^{(t)}$ is estimate of final weight vector after convergence of iterations at node i, we have:\\
\begin{align}
\begin{split}
f(\displaystyle\frac{\bar{\mathbf{w}}_i}{T})-f(\mathbf{w}^{\star})\leq\frac{2c}{\sqrt{\lambda}}+\frac{c^2 log(T)}{2T\lambda}+\frac{2}{\sqrt{\lambda}}\bigg(\frac{\gamma R}{\sqrt{\lambda}}+\gamma R\bigg)
\end{split}
\end{align}

\hfill$\square$

\section{Empirical Results}

In this section, we present the results on the data sets taking into consideration the time taken to load the data sets. We compare the time taken by the distributed algorithm to converge on all nodes against a centralized execution on a single server by estimating \emph{Speed-up}. Speed-up is defined as follows: 
\begin{equation}
\text{Speed-up} = \frac{\text{Time taken by the distributed algorithm to converge on all nodes}}{\text{Time taken by the centralized algorithm to converge}} 
\end{equation}
Our results indicate that in four of the seven datasets, GADGET outperforms the centralized algorithm with regard to time taken to load the data and build the model. In the remaining three datasets, the centralized algorithm is $1-3$ times faster than the distributed algorithm. These three datasets are dense and have a large number of features. Our experiments reveal that GADGET outperforms the centralized algorithm when the number of instances is significantly larger than the number of features.

\begin{table}[!h]
\begin{center}
\begin{tabular}{|c|cc|cc|c|} \hline
Dataset & \multicolumn{2}{l|}{GADGET}& \multicolumn{2}{l|}{Pegasos (Centralized)} & Speedup \\ \hline
& Time (secs) & (Acc. $\%$) & Time (secs) & (Acc.$\%$)  & Factor\\ \hline
Adult & 19.63 ($\pm 2.00$) & 77.04 ($\pm 0.03$) & 27.34 ($\pm 0.002$) & 68.79 ($\pm 0.18$)  & 0.72 \\ \hline
CCAT & 72.35 ($\pm 8.96$) & 84.99 ($\pm 0.016$) & 76.54 ($\pm 0.031$)  & 76.21 ($\pm 0.04$) & 0.94 \\ \hline
MNIST & 23.61 ($\pm 8.1$) & 88.57 ($\pm 0.04$) & 28.22 ($\pm 5.2$) & 89.81 ($\pm 0.01$)& 0.84 \\ \hline
Reuters & 15.45 ($\pm 0.81$) & 94.04 ($\pm 0.07$) & 11.07 ($\pm 6.04$) & 95.59 ($\pm 0.01$)&  1.39\\ \hline
USPS & 4.26 ($\pm 0.914$) & 92.12 ($\pm 0.03$) &6.59 ($\pm 2.05$) & 92.33 ($\pm 0.02$) & 0.65\\ \hline
Webspam & 23.21 ($\pm 4.21$) & 77.49 ($\pm 0.05$) & 21.09 ($\pm 2.2$) & 80.03 ($\pm 0.0$) & 1.10\\ \hline
Gisette & 204.68 ($\pm 135.84$) & 55.43($\pm 0.04$) & 71.50 ($\pm 32.04$) & 50.0($\pm 0.04$) & 2.86 \\ \hline
\end{tabular}
\end{center}
\caption{Comparison of the performance of GADGET SVM and centralized Pegasos on five data sets using the following metrics: accuracy of classification, time taken for the distributed algorithm versus the centralized including time to load data, and percentage of speed-up in execution time. The accuracy and time reported for GADGET is the mean over all the nodes in the network with the corresponding standard deviations. Epsilon at convergence $\epsilon$ = 0.00307, 0.00011, 0.00065, 0.00033, 0.01152 for the Adult, CCAT, MNIST, Reuters and USPS datasets respectively. }
\label{fig:tbPerfLoad}
\end{table}%

 \end{APPENDIX}

\end{document}